\documentclass{article}
\usepackage{spconf, amsmath, graphicx, multirow}

\def\L{{\cal L}}

\title{Fixed-Point Performance Analysis of Recurrent Neural Networks}
\name{Sungho Shin, Kyuyeon Hwang, and Wonyong Sung\thanks{This work was supported in part by the Brain Korea 21 Plus Project and the National Research Foundation of Korea (NRF) grant funded by the Korea government (MSIP) (No. 2015R1A2A1A10056051).}}
\address{Department of Electrical and Computer Engineering\\ Seoul National University\\ Seoul, 08826 Korea\\ Email : shshin@dsp.snu.ac.kr, kyuyeon.hwang@gmail.com, wysung@snu.ac.kr }
\begin{document}
%
\maketitle
\begin{abstract}
Recurrent neural networks have shown excellent performance in many applications; however they require increased complexity in hardware or software based implementations. The hardware complexity can be much lowered by minimizing the word-length of weights and signals. This work analyzes the fixed-point performance of recurrent neural networks using a retrain based quantization method. The quantization sensitivity of each layer in RNNs is studied, and the overall fixed-point optimization results minimizing the capacity of weights while not sacrificing the performance are presented. A language model and a phoneme recognition examples are used.
\end{abstract}
\section{Introduction}
\label{sec:introduction}

\indent Recurrent neural networks (RNNs) employ feedback paths inside, and they are suitable for processing input data whose dimension is not fixed. Important applications of RNNs include language models for automatic speech recognition, human action recognition and text generation~\cite{mikolov2011extensions, baccouche2011sequential, vinyals2014show}. 
\begin{figure}[t]
\begin{minipage}[]{1.0\linewidth}
  \centering
  \centerline{\includegraphics[width=7.8cm]{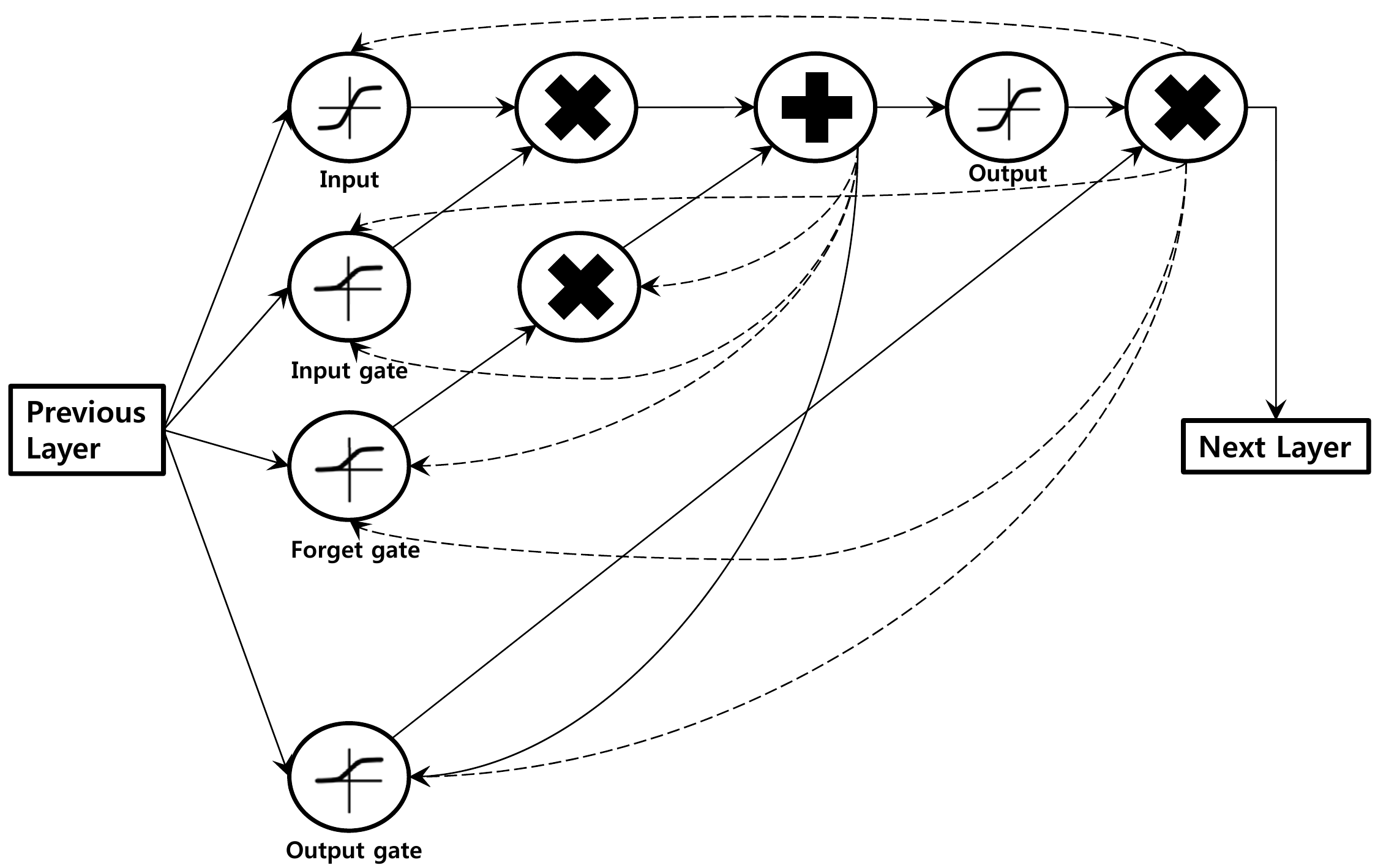}}
  \centerline{(a) Each layer of LSTM RNN}\medskip
\label{fig:lstm_a}
\end{minipage}
\begin{minipage}[]{1.0\linewidth}
  \centering
  \centerline{\includegraphics[width=8.5cm]{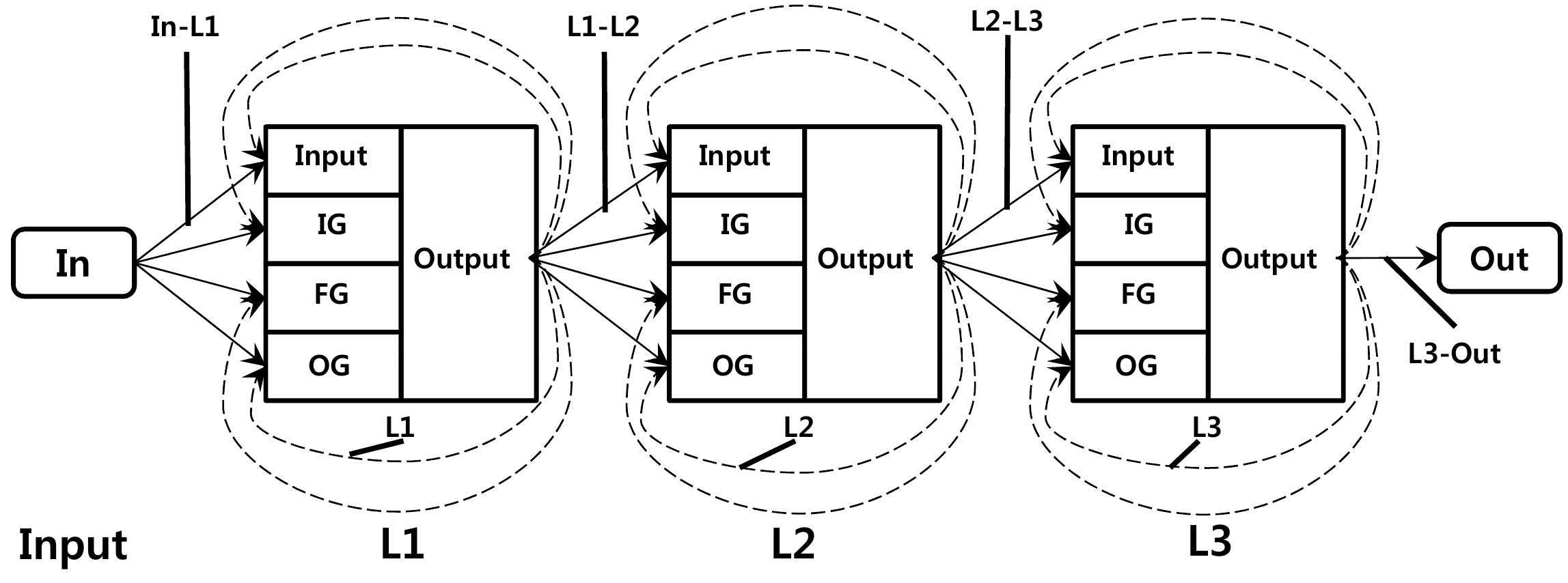}}
  \centerline{(b) Whole connection of three layer LSTM RNN}\medskip
\label{fig:lstm_b}
\end{minipage}
\caption{The standard structure of LSTM RNN.  Each circle represents one layer which is part of the LSTM. A dotted line means a backward path and a solid line is a forward path. The plus and multiplication signs show aggregation functions for summing and multiplication, respectively. The graph in the circles means an activation function for logistic sigmoid or tanh.}
\label{fig:lstm}
\end{figure}
\\
\indent Among many types of  RNNs~\cite{elman1990finding, hochreiter1997long, jaeger2001echo}, the most powerful one is the long short term memory (LSTM) RNN~\cite{hochreiter1997long}. A standard LSTM RNN is depicted in \figurename~\ref{fig:lstm}. A layer of LSTM RNN contains an input gate, an output gate, and a forget gate. An LSTM layer with $N$ units demands a total of approximately $4N^{2}+4NM+5N$ weights where $M$ is the unit size of the previous layer. Considering a character level language model that employs three 1024 size LSTM layers, the network demands about 22.3 million weights. Thus, reducing the size of weights in RNNs is very important for VLSI or embedded computer based implementations.
\\
\indent There have been many studies on efficient hardware implementation of artificial neural networks applying direct quantization~\cite{ holi1993finite, dundar1994effects, farabet2010hardware,tang1993multilayer}. Retraining based quantization that readjust the fixed-point weights by training after direct quantization of floating-point parameters was adopted in~\cite{kim2014x1000, hwang2014fixed, anwar2015fixed}. Previous research works for retrain-based fixed-point optimization use 3-8 bits for implementing  feedforward neural networks (FFNNs) and convolutional neural networks (CNNs). However RNN (especially LSTM RNN) employs a more complex feed-back based structure and hence optimum quantization is challenging.
\\
\indent In this paper, we optimize the word-length of weights and signals for fixed-point LSTM RNNs using the retraining method. The proposed scheme consists of three parts which are floating point training, sensitivity analysis of fixed-point RNNs and retraining. To the best of our knowledge, this is the first work that tries fixed-point quantization of large size recurrent neural networks.
\\
\indent This paper is organized as follows. In Section~\ref{sec:quantization}, the quantization procedure for weights and signals is  given. Section~\ref{sec:sensitive} describes layerwise fixed-point sensitivity analysis for weights and signals. Experimental results are provided in Section~\ref{sec:experiment} and concluding remarks follow in Section~\ref{sec:conclusion}.

\section{Retrain-based weight quantization}
\label{sec:quantization}
The retrain based quantization method includes the fixed-point conversion process inside of the training procedure so that the network learns the quantization effects~\cite{hwang2014fixed}. This method shows much better performance when the number of bits is small. In this method, the floating-point weights for RNNs are prepared with the backpropagation through time (BPTT) algorithm~\cite{werbos1990backpropagation}. BPTT begins by unfolding a recurrent neural network through time, then training proceeds in a manner similar to adapting a feed-forward neural network with backpropagation. 
\\
\indent  For direct quantization, a uniform quantization function is employed and the function $Q(\cdot)$ is defined as follows:
\begin{align}
	Q(w)=& sgn(w) \cdot \Delta  \cdot min\biggl(\left \lfloor{\dfrac{\lvert w \rvert}\Delta +0.5}\right \rfloor ,\dfrac{M-1}2\biggr) = \Delta \cdot z,  \label{eq:uniform_quant}
\end{align}  
where $sgn(\cdot)$ is the sign function, $\Delta$ is a quantization step size, $w$ is the set of the floating-point weights, and $M$ represents the number of quantization levels. Note that $M$ is normally an odd number since the weight values can be positive or negative. When $M$ is 5, the weights are represented by $-2\Delta$,  $-\Delta$, $0$, $\Delta$, and $2\Delta$, which can be stored in 3 bits. 
\\
\indent For selecting a proper step size $\Delta$, the L2 error minimization criteria is applied as adopted in~\cite{hwang2014fixed}. The quantization error $E$ is represented as follows:
\begin{align}
	E = \dfrac1 2\sum\limits_{i=1}^N\bigl(Q(w_{i})-w_{i} \bigr)^2 = \dfrac1 2\sum\limits_{i=1}^N\bigl(\Delta \cdot z_{i}-w_{i} \bigr)^2,   \label{eq:error}
\end{align}
where $N$ is the number of weights in each layer, $w_{i}$ is the $i$-th weight value in floating-point, and $z_{i}$ is the integer membership of $w_{i}$. 
The quantization error $E$ is minimized by the following two step iterative computation.
\begin{align}
	\boldsymbol{z}^{(t)} =&  \operatornamewithlimits{argmin}\limits_{{\boldsymbol{z}}}\operatorname{\textit{E}}(\boldsymbol{\boldsymbol{w}}, \boldsymbol{z}, \Delta^{(t-1)}) \nonumber \\ =& sgn(w_{i}) \cdot min\biggl(\left \lfloor{\dfrac{\lvert w_{i} \rvert}{\Delta^{(t-1)}} +0.5}\right \rfloor, \dfrac{M-1}2\biggr)\label{eq:solve1}   \\
	\Delta^{(t)} =&  \operatornamewithlimits{argmin}\limits_{{\Delta}}\operatorname{\textit{E}}(\boldsymbol{w}, \boldsymbol{z}^{(t)}, \Delta) = \dfrac {\sum\limits_{i=1}^N w_{i} \cdot z^{(t)}_{i} }{\sum\limits_{i=1}^N\bigl(z_{i}^{(t)} \bigr)^2},   \label{eq:solve2}
\end{align}
where the superscript ($t$) indicates the iteration step. The first step equation (\ref{eq:solve1}) can be computed using (\ref{eq:uniform_quant}). The second step equation (\ref{eq:solve2}) can be solved by using the derivative of the error with respect to $\Delta^{(t)}$ to be zero. The iteration stops when $\Delta^{(t)}$ is converged. 
\\
\indent After obtaining the fixed-point weights, $z_{i}\Delta$, the retraining procedure follows. We maintain both floating-point and quantized weights, since applying BPTT algorithm directly with quantized weights usually does not work. The reason is the amount of weights to be changed on each training step is much smaller than the quantization step size $\Delta$. Assuming that the RNN is unfolded, the algorithm can be described as follows:
\begin{gather}
net_{i} = \sum\limits_{j \in A_{i}}w_{ij}^{(q)}y_{j}^{(q)} \nonumber \\
y_{i}^{(q)}=R_{i}(\phi_{i}(net_{i})) \label{eq:forward} \\ \nonumber \\
\delta_{j}=\phi_{j}^{'}(net_{j})\sum\limits_{i \in P_{j}}\delta_{i}w_{ij}^{(q)}  \label{eq:backward}\\
\dfrac {\partial E}{\partial w_{ij}} = -\delta_{i}y_{j}^{(q)} \label{eq:gradient} \\
w_{ij, new} = w_{ij} - \alpha \biggl\langle \dfrac {\partial E}{\partial w_{ij}} \biggr\rangle \nonumber \\
w_{ij, new}^{(q)} = Q_{ij}(w_{ij, new}) \label{eq:weight}
\end{gather}
where $net_{i}$ is the summed input value of the unit $i$, $\delta_{i}$ is the error signal of the unit $i$, $w_{ij}$ is the weight from the unit $j$ to the unit $i$, $y_{j}$ is the output signal of the unit $j$, $\alpha$ is the learning rate, $A_{i}$ is the set of units anterior to the unit $i$, $P_{j}$ is the set of units posterior to the unit $j$, $R(\cdot)$ is the signal quantizer, $Q(\cdot)$ is the weight quantizer, $\phi(\cdot)$ is the activation function, the superscript $(q)$ indicates quantization, and $\langle\cdot\rangle$ is an average operation via the mini-batch. Equation (\ref{eq:forward}), (\ref{eq:backward}), (\ref{eq:gradient}), and (\ref{eq:weight}) represent the forward, backward, gradient calculation, and weights update phases each.
\section{Quantization sensitivity analysis}
\label{sec:sensitive}
LSTM RNNs usually contain millions of weights and thousands of signals. Therefore, it is necessary to group them according to their range and the quantization sensitivity~\cite{sung1995simulation}. Fortunately, a neural network can easily be layerwisely grouped. Throughout this sensitivity check, we can identify which layer in the neural network needs more bits for quantization. The network for phoneme recognition contains three hidden RNN layers. Thus the weights can be grouped into 7 groups, which are In-L1, L1, L1-L2, L2, L2-L3, L3, and L3-Out groups, where In-L1 connects input and the first LSTM layer and L1 is the recurrent path in the first level LSTM. ~\figurename\ref{fig:lstm}.(b) illustrates the grouping. In the sensitivity analysis, we only quantize the weights of the selected group while those in other groups are unquantized.
\begin{figure}[t]
  \centering
  \centerline{\includegraphics[width=8.5cm]{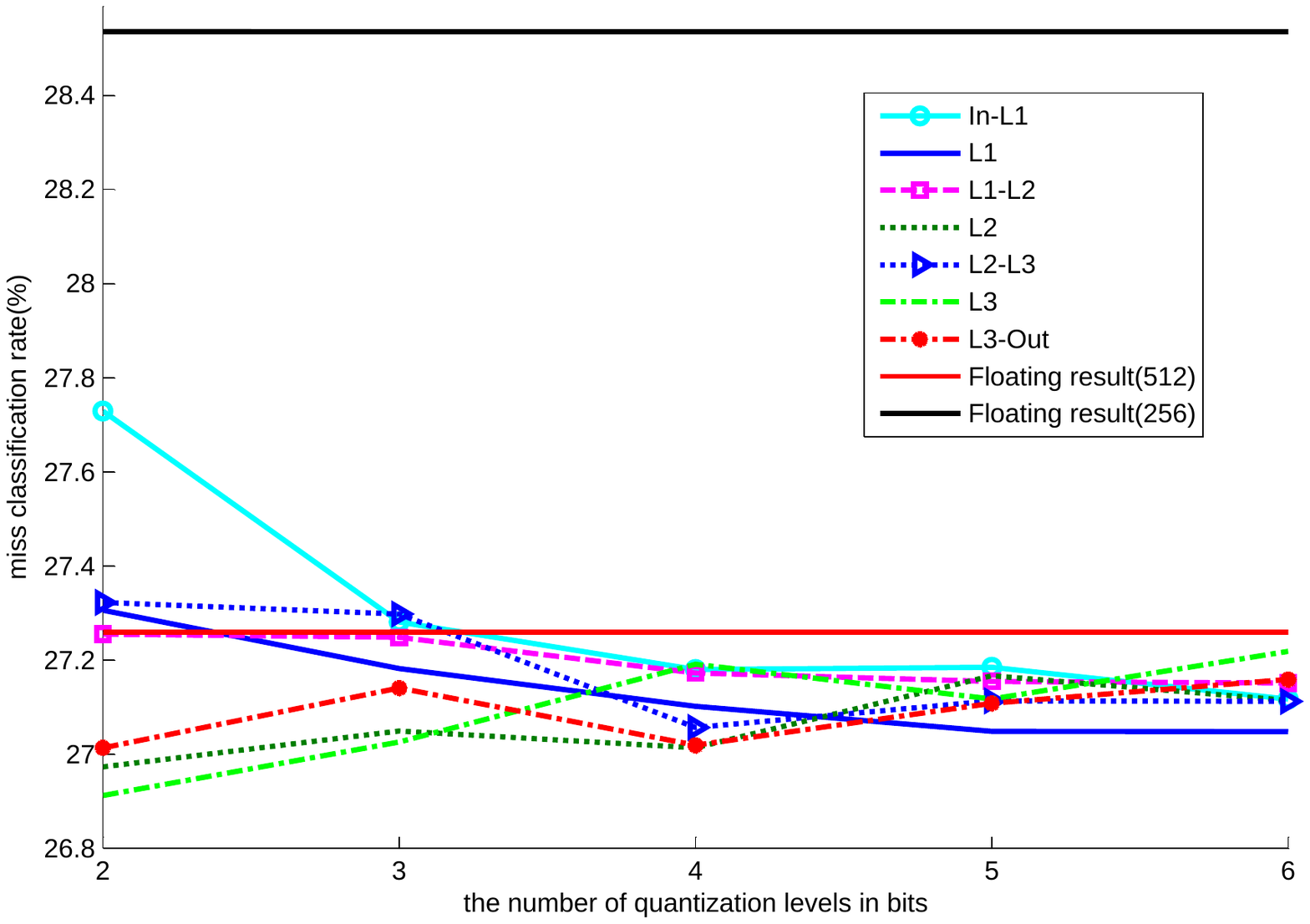}} 
\caption{Layerwise sensitivity analysis results of the weights in the phoneme recognition example. The red and black horizontal lines indicate the floating-point results for 512 LSTM size and 256 LSTM size each.}
\label{fig:phn_sensitive_analysis}
\end{figure}
Signal layerwise sensitivity analysis also follows the same scheme. In the standard LSTM RNN, the popular activation functions are logistic sigmoid or tanh. 
\begin{align}
	sigmoid(x) = \dfrac 1 {1+e^{-x}}   \label{eq:sigmoid} \\
	tanh(x) = \dfrac {e^{x}+e^{-x}}{e^{x}-e^{-x}}   \label{eq:tanh}
\end{align}
The output ranges of the sigmoid and the tanh are limited by 0 to 1 and -1 to 1, respectively. The quantization step size $\Delta$ is determined by the quantization level $M$. For example, if the signal word-length is two bits ($M$ is four), the quantization points are 0/3, 1/3, 2/3, and 3/3 for the sigmoid and -1/1, 0/1, and 1/1 for the tanh. However signals of linear units are not bounded and their quantization range should be determined empirically. In our phoneme recognition example, each component of the input data is normalized to have zero mean and a unit variance over the training set. The input range is chosen to be from -3 to 3. One hot encoding is used for the input linear units in the language model example.
\begin{table}[t]
\centering
\caption{Frame-level phoneme error rates (\%) on the test set with the TIMIT phoneme recognition example using quantized weights and signals. Numbers in the parenthesis indicate the ratio of the weights capacity compared to the floating-point version.}
\label{table 1}
\begin{tabular}{cccc|l|l|}
\hline
\multicolumn{2}{c}{Layerwise quantization bits}                                                                                                    & \multicolumn{4}{c}{FER(\%)}       \\ \hline
\begin{tabular}[c]{@{}c@{}}Weights bits\\(In-L1, L1, L1-L2, L2, \\L2-L3, L3, L3-Out)\end{tabular} & \begin{tabular}[c]{@{}c@{}}Signal bits\\ (Input, L1, \\L2, L3)\end{tabular} & Direct       & \multicolumn{3}{c}{Retrain} \\ \hline\hline
3-2-2-2-2-2-2 (6.39\%)                                                                             & 4-4-3-5                                                              & 48.00        & \multicolumn{3}{c}{28.74}   \\ \hline
4-3-3-3-3-3-3 (9.52\%)                                                                            & 4-4-3-5                                                              & 34.37        & \multicolumn{3}{c}{28.87}   \\ \hline
3-2-2-2-2-2-2 (6.39\%)                                                                             & 5-5-4-6                                                              & 31.65        & \multicolumn{3}{c}{27.79}   \\ \hline
4-3-3-3-3-3-3 (9.52\%)                                                                             & 5-5-4-6                                                              & 31.54        & \multicolumn{3}{c}{27.74}   \\ \hline
\end{tabular}
\end{table}
\section{Experimental Results}
\label{sec:experiment}
In this section, we will first show the result of the layerwise sensitivity analysis for signals and weights, and then fully quantized network performances. The proposed quantization strategy is evaluated using two RNNs, one for  phoneme recognition and the other for character level language modeling. Advanced training techniques such as early stopping, adaptive learning rate, and Nesterov momentum are employed~\cite{giles2001overfitting,weir1991method, nesterov1983method}. 
\subsection{Phoneme Recognition}
\label{ssec:experiment_phone}
 Phoneme recognition experiments were performed on the TIMIT corpus~\cite{garofolo1993darpa}. The detailed experimental conditions are the same with~\cite{graves2013speech} except the mapping of output classes for scoring. Therefore, the input layer consists of 123 linear units (Fourier-transfomed-based filter-bank with 40 coefficients plus energy with their first and second temporal derivatives). Three hidden LSTM layers have the same size of 512. The output layer consists of 61 softmax units which correspond to 61 target phoneme labels. The network is trained using the Fractal with the training parameters of 32 forward steps and 64 backward steps with 64 streams~\cite{hwang2015single}. Initial learning rate was $10^{-5}$ and it is decreased at proper timing until $10^{-7}$. Momentum was 0.9~\cite{jacobs1988increased} and adadelta was adopted for weights update~\cite{zeiler2012adadelta}. The network demands approximately 5.5 million weights. As a result it needs about 22MB with a 32bit floating-point format.
\\
\indent \figurename~\ref{fig:phn_sensitive_analysis} shows the result of the layerwise sensitivity analysis for the weights. The original phoneme frame error rate was 27.26\% with the LSTM layer size of 512, and that with the LSTM size of 256 was 28.63\%. The result indicates that all layers except the input-LSTM1 group shows the almost the same quantization sensitivity and requires only two bits for weight representation. Input-LSTM1 weights group demands at least three quantization bits. In the signal sensitivity analysis, all layers need only three to five quantization bits.  
\\
\indent Using the sensitivity analysis results, we construct a fully quantized LSTM RNN and the results are shown in ~\tablename\ref{table 1}. The result shows that the phoneme error rate of 27.74\% is achieved with only about 10\% of the weight capacity needed for the floating-point version.
\begin{figure}[tb]
  \centering
  \centerline{\includegraphics[width=8.5cm]{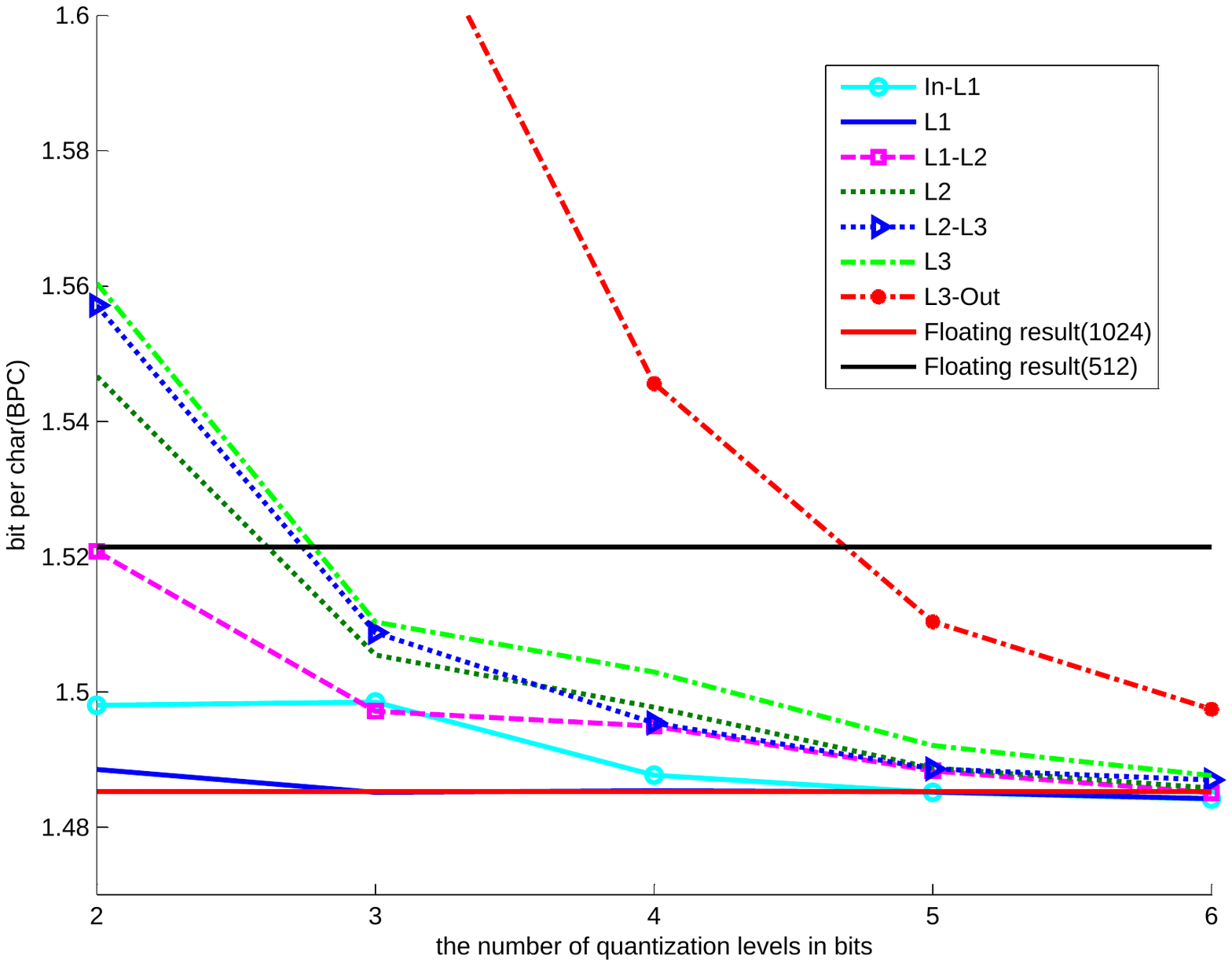}}
\caption{Layerwise sensitivity analysis results of the weights in the language model example. The red and black horizontal lines indicate the floating-point results for 1024 LSTM size and 512 LSTM size each.}
\label{fig:language_sensitive_analysis}
\end{figure}
\begin{table}[tb]
\centering
\caption{Bit per character on the test set with the English Wikipedia language model example using quantized weights and signals.  Numbers in the parenthesis indicate the ratio of the weights capacity compared to the floating-point version.}
\label{table 2}
\begin{tabular}{cccc|l|l|}
\hline
\multicolumn{2}{c}{Layerwise quantization bits}                                                                                                    & \multicolumn{4}{c}{BPC}       \\ \hline
\begin{tabular}[c]{@{}c@{}}Weights bits\\ (In-L1, L1, L1-L2, L2, \\L2-L3, L3, L3-Out)\end{tabular} & \begin{tabular}[c]{@{}c@{}}Signal bits\\ (L1, L2, L3)\end{tabular} & Direct       & \multicolumn{3}{c}{Retrain} \\ \hline\hline
2-2-3-4-4-4-6 (10.52\%)                                                                            & 6-6-7                                                             & 3.623        & \multicolumn{3}{c}{1.546}   \\ \hline
3-3-4-5-5-5-7 (13.64\%)                                                                            & 6-6-7                                                             & 1.641       & \multicolumn{3}{c}{1.510}   \\ \hline
4-4-5-6-6-6-8 (16.75\%)                                                                             & 6-6-7                                                              & 1.517        & \multicolumn{3}{c}{1.499}   \\ \hline
2-2-3-4-4-4-6 (10.52\%)                                                                             & 7-7-8                                                              & 3.613        & \multicolumn{3}{c}{1.545}   \\ \hline
3-3-4-5-5-5-7 (13.64\%)                                                                             & 7-7-8                                                              & 1.639        & \multicolumn{3}{c}{1.508}   \\ \hline
4-4-5-6-6-6-8 (16.75\%)                                                                            & 7-7-8                                                              & 1.517        & \multicolumn{3}{c}{1.499}   \\ \hline
\end{tabular}
\end{table}
\subsection{Language Model}
\label{ssec:experiment_language}
A character level language model was trained using English Wikipedia dataset which was used in ~\cite{sutskever2011generating}.  Each character was put into the neural network input using their own ASCII code values with one-hot encoding, which needs 256 units.  
\\
\indent The input layer contains 256 linear units to accommodate real valued inputs. Three hidden LSTM layers have the same size of 1024. The output layer consists of 256 softmax units that correspond to 256 character level one-hot encoding. The network is trained using Fractal with the training parameters of 128 forward steps and 128 backward steps employing 64 streams~\cite{hwang2015single}. The learning rate was decreased from  $10^{-6}$ to $10^{-8}$ during training. Momentum was 0.9 and adadelta was adopted for weights update. This network needs approximately 22.3 million weights.
\\
\indent \figurename~\ref{fig:language_sensitive_analysis} shows the layerwise sensitivity analysis results for the weights. The bit per character (BPC) of the floating-point language model with the layer size of 1024 was 1.485. The analysis shows that the most sensitive weights group of the network is the L3-Out layer. Note that the last RNN layer is connected to the softmax layer. The first weights group shows low sensitivity when compared to the phoneme recognition example. This is because one-hot encoding of the ASCII code is used as for the input in this language model. The network has two types of paths, the forward and the recurrent connections. For both paths, the layer that is close to the output shows higher quantization sensitivity. The sensitivity analysis of signals shows the minimum of four or five bits for quantization. 
\\
\indent We next try fixed-point optimization of all signals and weights. While the sensitivity analysis quantizes only one group of weights or signals, the fixed-point optimization quantizes all the weights and signals simultaneously to find out the most optimum set of word-lengths. The results are summarized in~\tablename~\ref{table 2}. Unlike the phoneme recognition example, this application needs more quantization bits over the results obtained from the sensitivity analysis. A reasonable result was achieved with two more bits for both weights and signals than the sensitivity analysis result. The memory space needed is 16.75\% when compared to the floating-point representation. 
\section{Concluding Remarks}
\label{sec:conclusion}
This work investigates the fixed-point characteristics of RNNs for phoneme recognition and language modeling.  The retrain-based fixed-point optimization greatly reduces the word-length of weights and signals. In the phoneme recognition example, most of the weights can be represented in 3 bits, while the language modeling needs 5 or 6 bits for obtaining near floating-point results.  By this optimization, the weights capacity needed can be reduced to only 10\% or 17\% of that required for floating-point implementations. The reduced weights and signals can lead to efficient hardware implementations or higher cache memory hit ratio in software based systems.



\bibliographystyle{IEEEbib}
\bibliography{refs}

\begin{thebibliography}{10}

\bibitem{mikolov2011extensions}
Tom{\'a}{\v{s}} Mikolov, Stefan Kombrink, Luk{\'a}{\v{s}} Burget, Jan~Honza
  {\v{C}}ernock{\`y}, and Sanjeev Khudanpur,
\newblock ``Extensions of recurrent neural network language model,''
\newblock in {\em Acoustics, Speech and Signal Processing (ICASSP), 2011 IEEE
  International Conference on}. IEEE, 2011, pp. 5528--5531.

\bibitem{baccouche2011sequential}
Moez Baccouche, Franck Mamalet, Christian Wolf, Christophe Garcia, and Atilla
  Baskurt,
\newblock ``Sequential deep learning for human action recognition,''
\newblock in {\em Human Behavior Understanding}, pp. 29--39. Springer, 2011.

\bibitem{vinyals2014show}
Oriol Vinyals, Alexander Toshev, Samy Bengio, and Dumitru Erhan,
\newblock ``Show and tell: A neural image caption generator,''
\newblock {\em arXiv preprint arXiv:1411.4555}, 2014.

\bibitem{elman1990finding}
Jeffrey~L Elman,
\newblock ``Finding structure in time,''
\newblock {\em Cognitive science}, vol. 14, no. 2, pp. 179--211, 1990.

\bibitem{hochreiter1997long}
Sepp Hochreiter and J{\"u}rgen Schmidhuber,
\newblock ``Long short-term memory,''
\newblock {\em Neural computation}, vol. 9, no. 8, pp. 1735--1780, 1997.

\bibitem{jaeger2001echo}
Herbert Jaeger,
\newblock ``The “echo state” approach to analysing and training recurrent
  neural networks-with an erratum note,''
\newblock {\em Bonn, Germany: German National Research Center for Information
  Technology GMD Technical Report}, vol. 148, pp. 34, 2001.

\bibitem{holi1993finite}
Jordan~L Holi and Jenq-Neng Hwang,
\newblock ``Finite precision error analysis of neural network hardware
  implementations,''
\newblock {\em Computers, IEEE Transactions on}, vol. 42, no. 3, pp. 281--290,
  1993.

\bibitem{dundar1994effects}
Gunhan Dundar and Kenneth Rose,
\newblock ``The effects of quantization on multilayer neural networks.,''
\newblock {\em IEEE transactions on neural networks, a publication of the IEEE
  Neural Networks Council}, vol. 6, no. 6, pp. 1446--1451, 1994.

\bibitem{farabet2010hardware}
Cl{\'e}ment Farabet, Berin Martini, Polina Akselrod, Sel{\c{c}}uk Talay, Yann
  LeCun, and Eugenio Culurciello,
\newblock ``Hardware accelerated convolutional neural networks for synthetic
  vision systems,''
\newblock in {\em Circuits and Systems (ISCAS), Proceedings of 2010 IEEE
  International Symposium on}. IEEE, 2010, pp. 257--260.

\bibitem{tang1993multilayer}
Chuan~Zhang Tang and Hon~Keung Kwan,
\newblock ``Multilayer feedforward neural networks with single powers-of-two
  weights,''
\newblock {\em Signal Processing, IEEE Transactions on}, vol. 41, no. 8, pp.
  2724--2727, 1993.

\bibitem{kim2014x1000}
Jonghong Kim, Kyuyeon Hwang, and Wonyong Sung,
\newblock ``X1000 real-time phoneme recognition vlsi using feed-forward deep
  neural networks,''
\newblock in {\em Acoustics, Speech and Signal Processing (ICASSP), 2014 IEEE
  International Conference on}. IEEE, 2014, pp. 7510--7514.

\bibitem{hwang2014fixed}
Kyuyeon Hwang and Wonyong Sung,
\newblock ``Fixed-point feedforward deep neural network design using weights
  +1, 0, and -1,''
\newblock in {\em Signal Processing Systems (SiPS), 2014 IEEE Workshop on}.
  IEEE, 2014, pp. 1--6.

\bibitem{anwar2015fixed}
Sajid Anwar, Kyuyeon Hwang, and Wonyong Sung,
\newblock ``Fixed point optimization of deep convolutional neural networks for
  object recognition,''
\newblock in {\em Acoustics, Speech and Signal Processing (ICASSP), 2015 IEEE
  International Conference on}. IEEE, 2015, pp. 1131--1135.

\bibitem{werbos1990backpropagation}
Paul~J Werbos,
\newblock ``Backpropagation through time: what it does and how to do it,''
\newblock {\em Proceedings of the IEEE}, vol. 78, no. 10, pp. 1550--1560, 1990.

\bibitem{sung1995simulation}
Wonyong Sung and Ki-II Kum,
\newblock ``Simulation-based word-length optimization method for fixed-point
  digital signal processing systems,''
\newblock {\em Signal Processing, IEEE Transactions on}, vol. 43, no. 12, pp.
  3087--3090, 1995.

\bibitem{giles2001overfitting}
Rich Caruana Steve Lawrence~Lee Giles,
\newblock ``Overfitting in neural nets: Backpropagation, conjugate gradient,
  and early stopping,''
\newblock in {\em Advances in Neural Information Processing Systems 13:
  Proceedings of the 2000 Conference}. MIT Press, 2001, vol.~13, p. 402.

\bibitem{weir1991method}
Michael~K Weir,
\newblock ``A method for self-determination of adaptive learning rates in back
  propagation,''
\newblock {\em Neural Networks}, vol. 4, no. 3, pp. 371--379, 1991.

\bibitem{nesterov1983method}
Yurii Nesterov,
\newblock ``A method for unconstrained convex minimization problem with the
  rate of convergence o (1/k2),''
\newblock in {\em Doklady an SSSR}, 1983, vol. 269, pp. 543--547.

\bibitem{garofolo1993darpa}
John~S Garofolo, Lori~F Lamel, William~M Fisher, Jonathon~G Fiscus, and David~S
  Pallett,
\newblock ``Darpa timit acoustic-phonetic continous speech corpus cd-rom. nist
  speech disc 1-1.1,''
\newblock {\em NASA STI/Recon Technical Report N}, vol. 93, pp. 27403, 1993.

\bibitem{graves2013speech}
Alan Graves, Abdel-rahman Mohamed, and Geoffrey Hinton,
\newblock ``Speech recognition with deep recurrent neural networks,''
\newblock in {\em Acoustics, Speech and Signal Processing (ICASSP), 2013 IEEE
  International Conference on}. IEEE, 2013, pp. 6645--6649.

\bibitem{hwang2015single}
Kyuyeon Hwang and Wonyong Sung,
\newblock ``Single stream parallelization of generalized {LSTM}-like {RNN}s on
  a {GPU},''
\newblock in {\em Acoustics, Speech and Signal Processing (ICASSP), 2015 IEEE
  International Conference on}. IEEE, 2015, pp. 1047--1051.

\bibitem{jacobs1988increased}
Robert~A Jacobs,
\newblock ``Increased rates of convergence through learning rate adaptation,''
\newblock {\em Neural networks}, vol. 1, no. 4, pp. 295--307, 1988.

\bibitem{zeiler2012adadelta}
Matthew~D Zeiler,
\newblock ``Adadelta: An adaptive learning rate method,''
\newblock {\em arXiv preprint arXiv:1212.5701}, 2012.

\bibitem{sutskever2011generating}
Ilya Sutskever, James Martens, and Geoffrey~E Hinton,
\newblock ``Generating text with recurrent neural networks,''
\newblock in {\em Proceedings of the 28th International Conference on Machine
  Learning (ICML-11)}, 2011, pp. 1017--1024.

\end{thebibliography}

\end{document}